%% file: main_untrack.tex
\documentclass[a4paper]{elsarticle}
\usepackage[a4paper, total={6.5in, 9in}]{geometry}

\usepackage{graphicx} 
\usepackage{caption}
\usepackage{subcaption}
\usepackage{amsmath}
\usepackage{amssymb}
\usepackage{xcolor}
\usepackage{multirow}
\usepackage{dsfont}
\usepackage{doi}

\UseRawInputEncoding

\begin{document}
\title{ICTSurF: Implicit Continuous-Time Survival Functions with Neural Networks}

\author[1,2]{Chanon Puttanawarut\corref{cor1}}
\ead{chanonp@protonmail.com}
\author[2]{Panu Looareesuwan}
\author[2]{Romen Samuel Wabina}
\author[2]{Prut Saowaprut}

\date{}
\cortext[cor1]{Corresponding author.}
\address[1]{Chakri Naruebodindra Medical Institute, Ramathibodi Hospital, Mahidol University, Samutprakarn, Thailand}
\address[2]{Department of Clinical Epidemiology and Biostatistics, Faculty of Medicine Ramathibodi Hospital, Mahidol University, Bangkok, Thailand}

\begin{abstract}
	Background and objective: Recently, the rise of models based on deep neural networks (DNNs) has demonstrated enhanced effectiveness in survival analysis. However, deep learning dealing with survival analysis usually required specific architecture or strict discretization scheme limiting the temporal precision of input and output values.
 
    Methods: This research introduces the Implicit Continuous-Time Survival Function (ICTSurF), built on a continuous-time survival model, and constructs survival distribution through implicit representation. As a result, our method is capable of accepting inputs in continuous-time space and producing survival probabilities in continuous-time space, independent of neural network architecture. 
    
    Results: Comparative evaluations against existing methods underscore the remarkable competitiveness of our proposed approach. Furthermore, we highlight the advantages of a flexible discretization scheme, showing improved performance with a lower number of discretizations compared to a rigid scheme.
    
    Conclusion: Our model exhibits competitive performance compared to existing methods, enhancing the flexibility of input data.
\end{abstract}
\maketitle

\section{Introduction}
\label{sec:introduction}
Survival analysis, also known as time-to-event analysis, aims at estimating the survival distributions of a specific event and time-of-interests. Typically, the estimation of survival probability involves modeling a relationship between covariates and time-to-event that is typically partially observed; e.g., it may not be possible to observe the event status of the same sample throughout the entire  observed time. This presents one of the key challenges in the field of survival analysis.

The conventional approaches commonly employed in survival analysis include the Cox Proportional Hazards (CPH) model, as proposed by Cox \cite{cox_regression_1972}. Although the CPH model is widely used, it is burdened by a substantial assumption of a consistent proportional hazard throughout the entire lifespan and a predetermined relationship between covariates. Other conventional methods, such as Weibull or Log-Normal distribution, also model a relationship between time and covariates based on a strong parametric assumption.

Recently, due to the success of deep neural network (DNN) based models, the majority of research in survival analysis has shifted towards models built on DNNs, demonstrating superior performance compared to conventional approaches. Recent studies have shown that the majority of the survival models are an extension of the conventional CPH model \cite{wiegrebe_deep_2023}. It calculates the hazard rate at time $t$ in the form of $h(t)=h_0(t)e^{g(x)}$, where $e^{g(x)}$ is a relative risk that depends on covariate vector $x$ such that $g( \cdot )$ is estimated by a neural network. Initial investigations of adopting neural networks in survival analysis date back to 1995 when Faraggi-Simon et al. \cite{faraggi_neural_1995} utilized neural networks to predict individual risk events based on the CPH model assumption. However, this pioneering study failed to show the advantages of using such methods over conventional CPH models. The deep learning-based model with CPH assumption was then improved in DeepSurv \cite{katzman_deepsurv_2018} using modern deep learning techniques, which shows superior performance over the conventional CPH model. However, this deep learning model still follows the proportional hazards assumption. To overcome this, several approaches have been developed using deep learning. Cox-Time \cite{kvamme_time--event_2019} calculates the relative risk function as $e^{g(x,t)}$ to model the interaction between $x$ and $t$. Therefore, Cox-Time is more flexible and robust than DeepSurv since it does not depend on the CPH assumption, but is still based on the partial likelihood function.

Existing studies have developed discrete-time survival models that try to overcome the proportional hazards assumption. These models allow the input data at a predefined specific time only. For instance, Deephit \cite{lee_deephit_2018} uses a discrete-time likelihood that constructs a probability mass function (PMF) of event time using softmax as an output layer of the model. This approach can overcome the CPH assumption by learning the PMF directly. Deephit has been designed to deal with competing risks by adding a sub-network for each specific risk. Moreover, it has been extended to DynamicDeephit \cite{lee_dynamic-deephit_2020} which can also learn the PMF but can handle time-varying input data. Nnet-survival \cite{gensheimer_scalable_2019} is also a neural network model based on a discrete-time survival function. Instead of learning the PMF like in Deephit, Nnet-survival learns the hazard value at a discrete time point, which also uses the hazard to construct a survival function. Due to the architecture of these models, the survival probability can only be calculated at predefined times \cite{lee_dynamic-deephit_2020, gensheimer_scalable_2019}. Consequently, it is assumed that the survival probability remains constant within the same interval of time or requires interpolation to obtain the survival probability at a precise time. On the other hand, the implicit survival function (ISF) approach, based on discrete-time survival models, can predict hazards at specific times using sample properties and time as input \cite{ling_learning_2023}. Unlike previous methods, ISF can output continuous-time survival functions since time is provided as model input.

Kvamme and Borgan proposed the piecewise constant hazard (PC-hazard) model for survival analysis \cite{kvamme_continuous_2021}. The model itself is similar to Nnet-survival where the model outputs hazard values at prespecified discrete times. However, the loss function of PC-hazard is based on a continuous-time survival model, unlike Nnet-survival. SurvTrace \cite{wang_survtrace_2022} uses the same loss function as PC-hazard but also add other objective function when training the model. Its network is based on a transformer architecture \cite{vaswani_attention_2017} instead of a multi-layer perceptron.  

Another approach in survival analysis is parametric models. DeepWeiSurv \cite{lauw_estimation_2020} is a parametric deep learning-based method for survival analysis, where it calculates the PDF of event time by a mixture of Weibull distribution through a neural network. The Deep Survival Machine (DSM) \cite{nagpal_deep_2021} also employs a neural network that estimates parameters using primitive distributions. However, unlike DeepWeiSurv, DSM experiments with two kinds of primitive distribution,
Weibull and Log-Normal. It then constructs a PDF of event time as a mixture distribution. DSM is also the first study that demonstrates the utility of a deep learning-based parametric model for competing risk. However, this model is primarily dependent on the chosen primitive distribution.

Here we propose a survival analysis model based on a DNN called \textit{Implicit Continuous-Time Survival Function (ICTSurF)}. Our model is built upon a continuous-time survival model and constructing survival distribution using implicit function. Unlike the previous method using continuous-time survival function which can only receive an input and output survival probability at a predefined discrete-time. Furthermore, unlike previous method using an implicit function, we incorporate continuous-time survival model to enhance flexibility in the choice of discretization schemes.

\begin{figure}[!tb]
     \centering
     \begin{subfigure}[b]{0.45\textwidth}
         \centering
         \includegraphics[width=\textwidth]{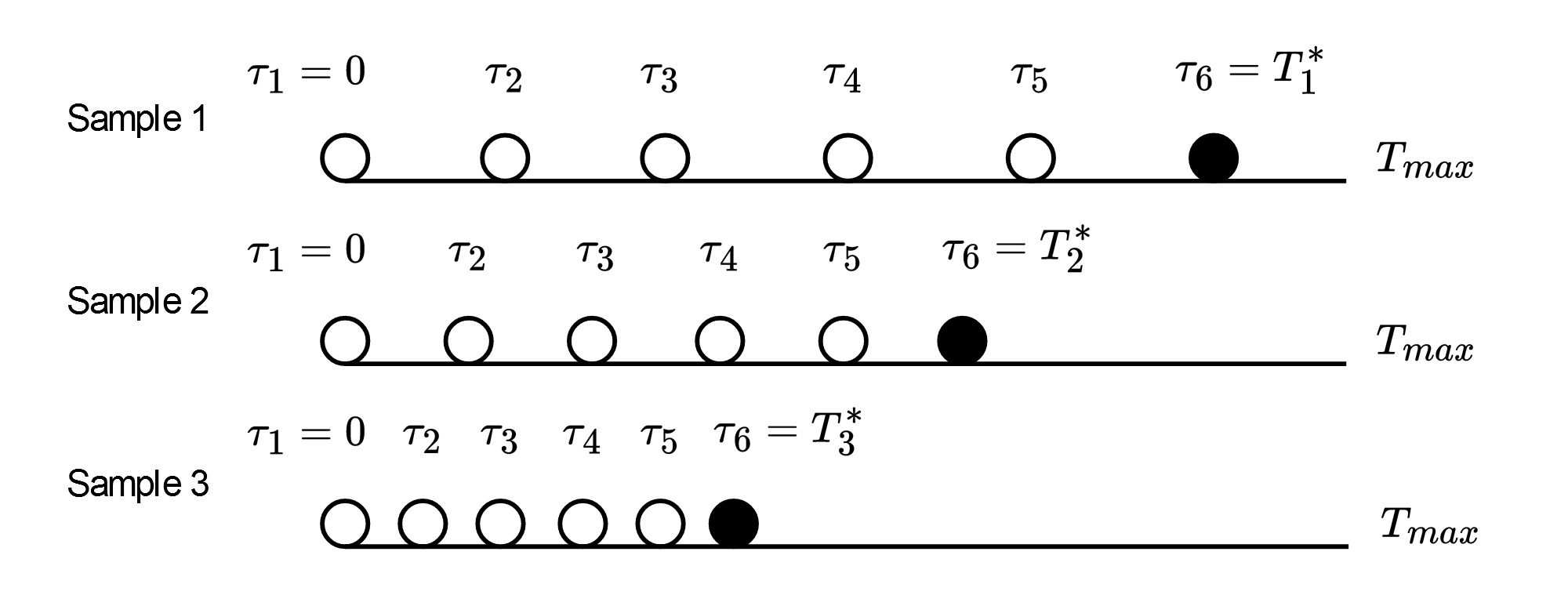}
         \caption{}
         \label{fig:cutequal}
     \end{subfigure}
     \hfill
     \begin{subfigure}[b]{0.45\textwidth}
         \centering
         \includegraphics[width=\textwidth]{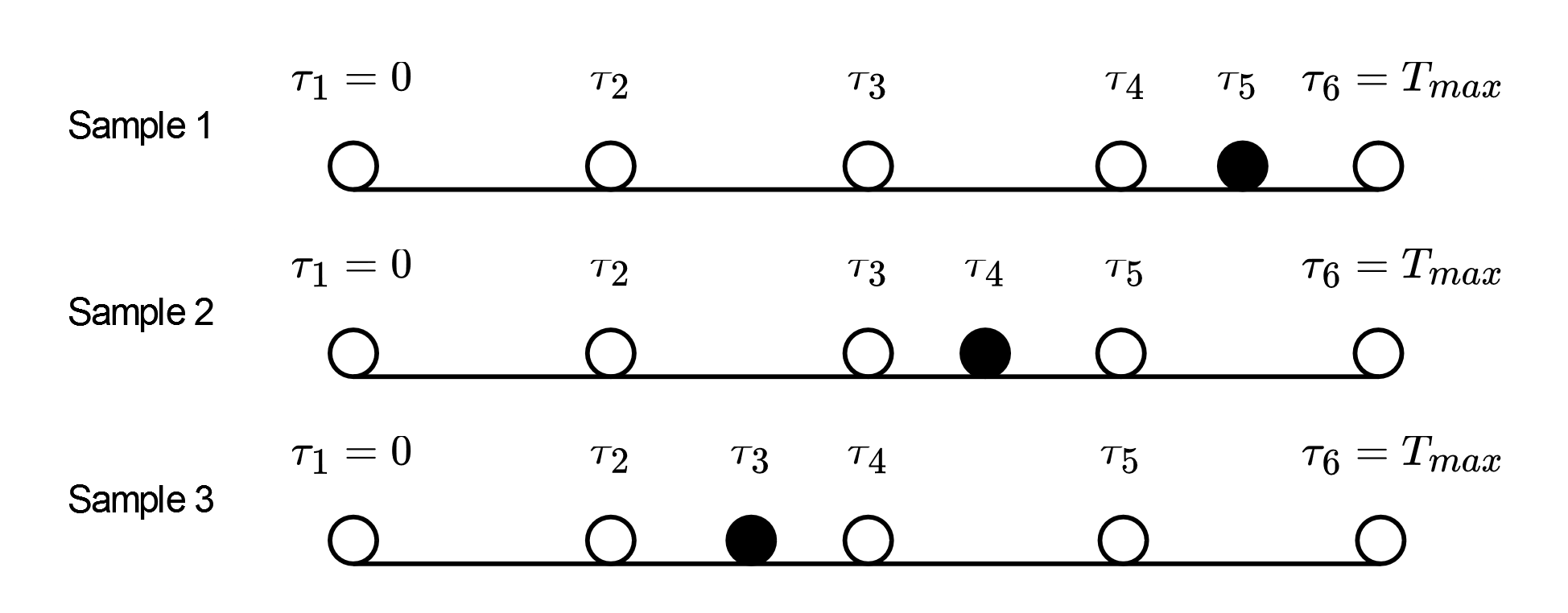}
         \caption{}
         \label{fig:cutsame}
     \end{subfigure}
     \caption{ Discretization schemes: Illustration of discretization schemes with black dots denoting time points corresponding to events ($T^*_i$). (a) Individual discretization is applied to each time point for every sample. (b) Attempted uniform discretization for all samples, with variations occurring only at the times of events.} 
     \label{fig:cuts}
\end{figure}

Our contribution can be summarized as
\begin{itemize}
    \item We introduced ICTSurF as a novel DNN model that calculates hazard rate at any given time point without requiring a fixed, predefined specific time for input data, thereby allowing the model to receive precise inputs.
    \item Our approach allows neural networks to learn nonlinear interaction between time and other covariates without a strong parametric assumption.
    \item We compare our model with the existing models using both the real-world and simulated datasets. The experiments were done in multiple datasets which include single risk and competing risk datasets.
    \item To explore the advantages of the continuous-time survival model, we examine two discretization approaches (refer to Figure \ref{fig:cuts}). The initial approach aims to leverage flexibility in discretization by treating each sample with distinct time point discretization (see Figure \ref{fig:cutequal}). The second approach aims for uniform discretization points across all samples (see Figure \ref{fig:cutsame}). A comparison between these two discretization schemes reveals that the flexibility in the discretization approach offers advantages over a more rigid scheme.

\end{itemize}

\section{Methods}

\subsection{Background}

In this section, we present fundamental concepts in continuous-time survival model.

Our dataset is composed of $n$ tuple of ${(x_i,T^*_i,k_i)}_{i=1}$, where $x_i \in \mathbb{R}^a$ represents an $a$-dimensional vector of covariates for sample $i$, $T^*_i$ represents the time of the event occurrence, and $k_i$ is event such that $ k_i \in \{ \text{\O}, 1, ..., K \}$, where $\text{\O}$ indicated right-censoring and $K$ is the total number of distinct events. We denote $d_i^k$ as $\mathds{1}(k_i \ne \text{\O})$, which $d_i^k = 1$ indicates an uncensored event, while $d_i^k = 0$ indicates a censored event. In this study, it is assumed that the censored data are right-censored, and the censor time is considered independent of the uncensored time.

The survival function $S(t)$ is a probability of the event that occurs for time greater than $t$, which can be expressed as:
$$S(t) = P(T^* > t)$$
where $T^*$ is the event time. Alternatively, $S(t)$ can be expressed in terms of the cumulative incidence of the event, $F(t)=P(T^*\le t)$, or the probability density function (PDF) of the event, and $f(t)$ denoted as $f(t) = P(T^*=t)$. The connection between $S(t)$, $F(t)$ and $f(t)$ is expressed through the equation $1-S(t) = F(t) = \int_0^t f(s)ds$. Note that for time with discrete value, the $f(t)$ uses the PMF of the event instead.

The hazard function, denoted as $h(t)$, is a fundamental component in survival analysis, defined as:
\begin{equation}
h(t)=\lim_{\delta \rightarrow 0} \frac{P(t \le T^* < t+\delta|T^*\ge t)}{\delta} = \frac{f(t)}{S(t)}.
\label{eq:hazard_function}
\end{equation}
This function represents the instantaneous rate at which the event occurs at time $t$, given no event has occurred up to that point. The survival function, $S(t)$, can then be expressed in terms of the hazard function as:
\begin{equation}
    S(t) = e^{-\int_0^t h(s) ds}
    \label{eq:surv_function}
\end{equation}
 
\subsection{Loss function}

Using Equation \ref{eq:hazard_function} and Equation \ref{eq:surv_function}, the likelihood that a sample $i$ undergoes event $k_i$ at time $t_i$ is expressible as follows:
\begin{equation}
\begin{split}
    L(t_i,d_i^k) &= P(T^* = t)^{d_i^k}P(T^* > t)^{1-d_i^k} \\
                &= f(t)^{d_i^k}S(t)^{1-d_i^k}\\
             &= {h(t_i)}^{d_i^k} S(t_i) = {h(t_i)}^{d_i^k} 
    e^{-\int_0^t h(s) ds}
\end{split}
\label{eq:likelihood1}
\end{equation}

Our proposed method is to parameterize the hazard function $h(t_i)$ directly using a deep neural network, denoted as $\hat{h}(t, x) \approx h(t)$. To optimize $\hat{h}(t,x)$, we have to maximize $L(T^*_i,d_i^k)$ by expressing \ref{eq:likelihood1} as 
\begin{equation}
L(T_i,d_i^k)=
\hat{h}(T^*_i,x)^{d_i^k}
 e^{-\int_0^{T_i} \hat{h}(s, x_i)ds }.
\label{eq:likelihood_nn}
\end{equation}

To compute the integral in Equation \ref{eq:likelihood_nn}, we discretize time of the sample $i$ into $m$ discrete time points, denoted as $0=\tau_{i,1}<\tau_{i,2}<...<\tau_{i,m}$. The spacing between consecutive times $\tau_{i,j}$ and $\tau_{i,j-1}$ for sample $i$ is denoted as $\Delta_{i,j}$, where $\Delta_{i,j} = \tau_{i,j} - \tau_{i,j-1}$. For simplicity, we choose $\Delta_{i,j}$ to be constant for all $j \in \{2, 3, ..., m\}$. To ensure that our network can receive input at specific times, we choose $\tau_{i,j}$ such that $T^*_i$ always corresponds to $\tau_m$. An illustration of this discretization scheme is provided in Figure \ref{fig:cutequal}. The integration was then approximated by numerical integration using the trapezoidal rule. The integral in Equation \ref{eq:likelihood_nn} can be approximated as:
\begin{equation}
    \int_0^{T^*_i} \hat{h}(s, x_i)ds \approx 
    \sum_{j=2}^m 
    \frac{\hat{h}(\tau_{i,j}, x_i)+\hat{h}(\tau_{i,j-1}, x_i)}{2}
    \times \Delta_{i,j}
\end{equation}

In summary, the optimization of the model for a single event ($k_i \in \{ \text{\O}, 1\}$) involves minimizing the average negative logarithm of the likelihood in Equation \ref{eq:likelihood_nn} across all samples, which can be written as:

\begin{equation}
\begin{split}
loss_{single} &= -\frac{1}{n}\sum_{i=1}^n \biggl[
d_i^k \log{\hat{h}(T^*_i,x)} \\ 
&-\sum_{j=2}^m 
    \frac{\hat{h}(\tau_{i,j}, x_i)+\hat{h}(\tau_{i,j-1}, x_i)}{2}
    \times \Delta_{i,j} \biggr].
\label{eq:likelihood_final}
\end{split}
\end{equation}

For managing competing risk datasets, the loss function treats a competing event as equivalent to right-censoring. The multi-event loss function is defined as follows:
\begin{equation}
\begin{split}
loss_{multi} &= \frac{1}{K} \sum_{k=1}^K \Biggl[ -\frac{1}{n}\sum_{i=1}^n \biggl[
\mathds{1}(k_i=k) \log{\hat{h}(T^*_i,x)} \\ 
&-\sum_{j=2}^m 
    \frac{\hat{h}(\tau_{i,j}, x_i)+\hat{h}(\tau_{i,j-1}, x_i)}{2}
    \times \Delta_{i,j} \biggr] \Biggr].
\label{eq:likelihood_final_multi}
\end{split}
\end{equation}

\subsection{Model architecture}

\begin{figure}[!tb]
     \centering
     \includegraphics[width=0.4\textwidth]{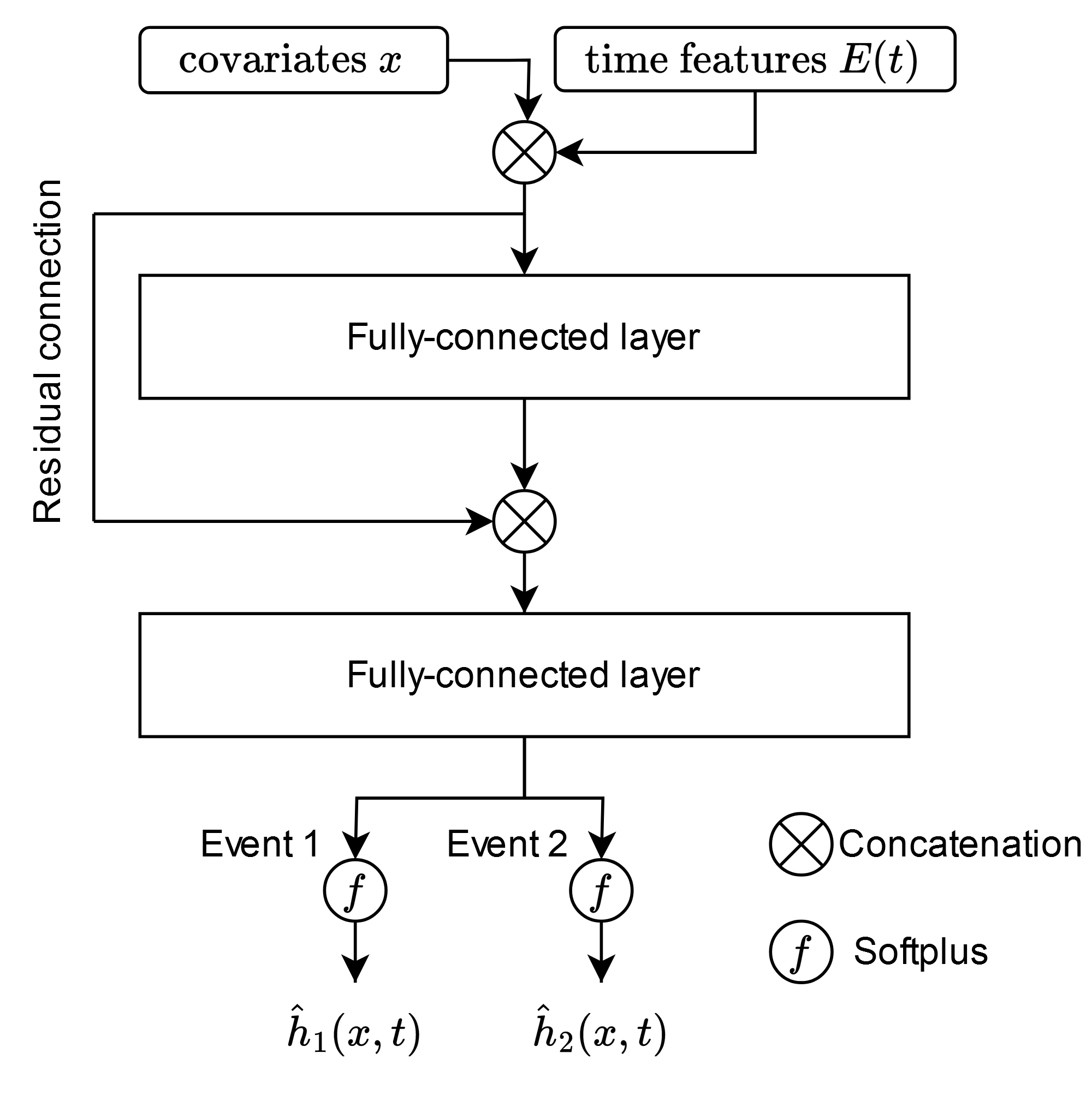}
     \caption{ Model architecture. The model takes covariates and a specific time of interest as input, producing hazard rates, denoted as $\hat{h}(x,t)$, for each event at the specified time as its output. Note that in this figure, we assume two competing risks, hence two hazard functions in the output. Conversely, in scenarios involving a single risk, there would be a single hazard function in the output.}
     \label{fig:arch}
\end{figure}

The overview of our model architecture is demonstrated in Figure \ref{fig:arch}. By any given covariate $x$ at time $t$, our model outputs an estimated hazard, $\hat{h}_k(t,x)$, for each event $k$. The input vector in the neural network is a vector of covariates $x$ concatenated with time features, denoted as $E(t)$. Here, $E(\cdot)$ represents the function that transforms time into time features. These time features, encapsulated by $E(t)$, can be obtained either through the embedding layer or by directly using raw time values (in which case $E(\cdot)$ acts as the identity function). To embed time, we employ the Time2Vec layer \cite{kazemi_time2vec_2019}, which is capable of capturing both periodic and non-periodic properties of time. The Time2Vec layer transforms the real value of time into an embedded vector of dimension $b$, where the non-periodic property is captured in the first element of the time-embedded vector. To capture periodic properties, the remaining elements undergo a periodic activation function, specifically the sine function, inspired by Positional Encoding \cite{vaswani_attention_2017}.

The input vector then passes through two fully connected layers. Within fully connected layers, each neuron conducts a linear transformation on the input vector via a weights matrix. Subsequently, a non-linear activation function (ReLU, as specified in section \ref{sec:experiment}) is applied to the resulting product. Before entering the second fully connected layer, the output vector from the first layer is concatenated with a residual connection from the input layer. The final hazard value, $\hat{h}(x, t)$, is then obtained using the Softplus activation function, where $\text{Softplus}(x) = \text{log}(1 + \text{exp}(x))$.

\section{Experiments}

Our experiments are divided into two parts. The first part uses previous models in survival analysis as the baseline while the second part compares two discretization schemes in regards to their flexibility advantage as continuous-time survival models. All experiments were implemented using PyTorch library \cite{paszke_pytorch_2019} in Python programming language.

\subsection{Datasets}

\input{dataset_untrack}

We categorized the datasets we utilized into two groups: single risk and competing risk datasets. For single risk, we use The Molecular Taxonomy of Breast Cancer International Consortium (METABRIC) \cite{curtis_genomic_2012}, The Study to Understand Prognoses and Preferences for Outcomes and Risks of Treatments (SUPPORT) \cite{knaus_support_1995} and DeepSurv's simulated survival data (SYNTHETIC NONLINEAR) \cite{katzman_deepsurv_2018}. As for competing-risk datasets, we created a synthetic dataset referred to as SYNTHETIC COMPETING. The details of the datasets are presented in Table \ref{tab:datasets}.

To prepare the datasets for modeling, numerical features across all datasets were normalized to zero mean and unit variance based on the training set. Categorical features underwent one-hot encoding before model training, except for the SurvTRACE model, which is designed to handle feature embedding. The time values were scaled using the mean time in the hold-out validation set.

\textbf{METABRIC:} The METABRIC dataset was originally built to determine breast cancer subgroups. There were 1,981 patients in the study, 888 (44.83\%) of them were followed until death and 1,093 (55.17\%) were right-censored. Variables in the dataset include gene expression and clinical characteristics. The dataset was downloaded from Deephit's Github\footnote{https://github.com/chl8856/DeepHit} which the dataset was already processed as stated in DeepHit paper \cite{lee_deephit_2018}.

\textbf{SUPPORT:} This is a multicenter study intended to investigate survival time for seriously ill hospitalized adult patients. The dataset has 9,105 patients in the study, 6,201 (68.11\%) were followed until death, and 2,904 (31.89\%) were right-censored. This dataset was downloaded from auton-survival's Github\footnote{https://github.com/autonlab/auton-survival}. We incorporated a total of 28 features into our analysis including Serum Albumin, PaO2/(.01*FiO2), Bilirubin, Serum Creatinine, BUN, White Blood Cell Count, Urine Output, Years of Education, SUPPORT Coma Score, Race, Respiration Rate, Serum Sodium, Glucose, ADL Surrogate, ADL Patient, Income, Average TISS, Mean Arterial Blood Pressure, Heart Rate, Temperature, Serum pH Arterial, Age, Gender, Disease Group, Disease Class, Diabetes, Dementia and Cancer. To handle missing values, we imputed the mean for numerical features and the mode for categorical features using all available data. However, for seven specific features: Serum Albumin, PaO2/(.01FiO2), Bilirubin, Serum Creatinine, BUN, White Blood Cell Count, and Urine Output, we employed fixed values for imputation as outlined in the original DSM paper \cite{nagpal_deep_2021}.

\textbf{SYNTHETIC NONLINEAR:} The synthetic dataset was obtained from DeepSurv's Github\footnote{https://github.com/jaredleekatzman/DeepSurv}. The dataset was generated by randomly selecting a 10-dimensional covariate vector $x$ from a uniform distribution $[-1,1)$. The first two covariates in $x$ were then utilized to construct a sample event time through an exponential Cox model \cite{austin_generating_2012} with a Gaussian log-risk function. Subsequently, the data was randomly censored to achieve a 50\% censoring rate. For further details on the simulation process, please refer to the original DeepSurv paper \cite{katzman_deepsurv_2018}.

\textbf{SYNTHETIC COMPETING:} We generated a synthetic dataset following a methodology similar to that described in \cite{ahmed_m_deep_2017}. Specifically, for each sample $i$:
$$\begin{aligned}
    X_i &\sim \mathcal{N}(0, I) \\
    T_i^1 &\sim \exp{\bigl(\text{cosh}(x_{i,1} + x_{i,2} + x_{i,3} + x_{i,4})\bigr)} \\
    T_i^2 &\sim \exp{\bigl(
    \big|\mathcal{N}(0, 1) + 
    \text{sinh}(x_{i,1} + x_{i,2} + x_{i,3} + x_{i,4})\big|
    \bigr)} \\
    T^*_i &= \text{min}(T_i^1, T_i^2) 
\end{aligned}$$

In this context, $X_i$ represents a 20-dimensional vector drawn independently from a normal distribution, and $x_{i,j}$ is $j$-th element in vector $X_i$ for the $i$-th sample. The simulation involves two event risks. The occurrence time for event 1, denoted as $T_i^1$, is sampled from an exponential distribution with a mean parameter dependent on the hyperbolic cosine ($cosh$) of the sum of the first four elements in vector $X_i$. The occurrence time for event 2, denoted as $T_i^2$, is sampled from an exponential distribution with a mean parameter dependent on the hyperbolic sine ($sinh$) of the sum of the first four elements in vector $X_i$ and a random variable drawn from a normal distribution. The final event time, $T^*_i$, is determined as the minimum of $T_i^1$ and $T_i^2$. A dataset of 30,000 samples is generated, after which 50\% of all samples are randomly selected to be right-censored.

\subsection{Baselines}

For performance comparison, we used CPH \cite{cox_regression_1972}, PC-Hazard \cite{kvamme_continuous_2021}, DeepHit \cite{lee_deephit_2018}, DSM \cite{nagpal_deep_2021} and SurvTRACE \cite{wang_survtrace_2022} as baselines for single-risk datasets. We used cause-specific CPH (cs-CPH), DeepHit and DSM as baselines for competing risk datasets.

The implementation of the PC-Hazard model was obtained from pycox's Github\footnote{https://github.com/havakv/pycox}, with the set of hyperparameter choices following the original paper \cite{kvamme_continuous_2021}. In the case of DeepHit, we also utilized the implementation from pycox's GitHub. The number of nodes and layers was held constant as outlined in the original paper \cite{lee_deephit_2018}. The selection of hyperparameter $\alpha$ was determined based on the set of hyperparameters implementation in DeepHit's GitHub, and $\sigma$ was fixed at a constant value of 0.1. For DSM, we employed the auton-survival package \cite{nagpal_auton-survival_2022}, with hyperparameters similar to the original DSM paper \cite{nagpal_deep_2021}. Regarding SurvTRACE, we used the implementation from SurvTRACE's GitHub\footnote{https://github.com/RyanWangZf/SurvTRACE}, and the choice of hyperparameters followed the specifications in the original SurvTRACE paper \cite{wang_survtrace_2022}.

\subsection{Metrics}

In evaluating the model's discriminative performance, we utilize time-dependent Concordance-Index ($C^{td}$-index) \cite{antolini_timedependent_2005} which is an extension of Harrell's Concordance-Index ($C$-index) \cite{harrell_evaluating_1982}. By integrating Inverse Probability Censoring Weighting (IPCW) with the concordance index, the censoring distribution is adjusted to correct the bias due to censoring. This metric enables us to assess the model's capability in differentiating the level of risk in the dataset \cite{uno_c-statistics_2011}.

We apply the time-dependent Brier score with IPCW \cite{graf_assessment_1999} to evaluate the model's calibration between the predicted survival probabilities and the actual outcome for the single risk task. 
Meanwhile, the standard concordance index and Brier score are applied since the true time of events for competing risks is known. These metrics allow us to ensure the model's efficacy in accurately ranking the risks and quantifying the predicted probabilities. Both metrics are evaluated at the 25\%, 50\%, and 75\% percentiles across the time horizon.

\subsection{Experiment with baselines}
\label{sec:experiment}

\begin{figure}[!tb]
     \centering
     \includegraphics[width=0.4\textwidth]{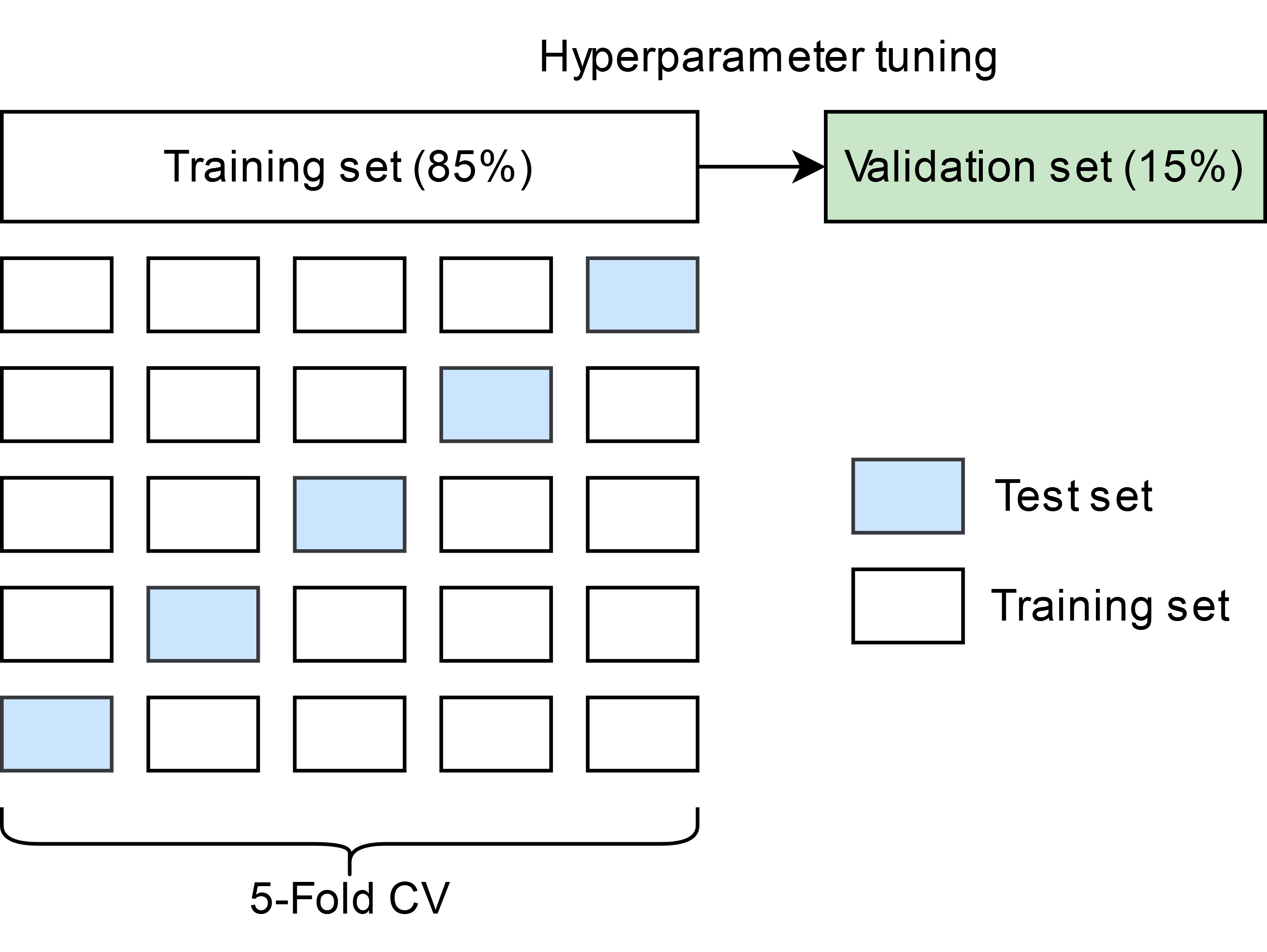}
     \caption{ 5-fold CV. Hyperparameters were first selected using a holdout validation set. Performance metrics were then evaluated using 5-fold CV.}
     \label{fig:cv}
\end{figure}

Our models were trained by using Adam \cite{kingma_adam_2017} as the optimizer with a fixed learning rate of $2\times10^{-4}$. ICTSurF was built with two hidden layers, where the number of nodes in the first and second hidden layers was chosen separately from \{64, 128\}. The dimensions of the embedded vector, $b$, were chosen from \{16, 32\}. After each hidden layer, batch normalization \cite{ioffe_batch_2015} was performed, followed by ReLU activation \cite{agarap_deep_2019}. The number of discretization time points was set to 50, while the batch size was set to 256. For hyperparameter tuning, the dataset was initially split into a holdout validation set (15\%). Hyperparameters were selected by minimizing the loss on this validation set. The remaining data underwent five-fold cross-validation (CV). The best models were selected based on the minimum loss on the holdout validation set. Model performance was evaluated by reporting the mean and standard error of metrics across all folds. This split was designed to prevent data in the validation set from leaking into the test set, which would produce data leakage.  Figure \ref{fig:cv} illustrates the 5-fold CV experiment.

Moreover, we additionally perform a time features extraction experiment. We contrast ICTSurF employing Time2Vec as the embedding layer with two other models: one that utilizes time without encoding ($\text{ICTSurF}_{\text{woe}}$) and another where ICTSurF incorporates Positional Encoding ($\text{ICTSurF}_{\text{pe}}$) \cite{vaswani_attention_2017}.

\subsection{Discretization experiment}
Here, we compare two discretization schemes in our ICTSurF model (Figure \ref{fig:cuts}). The initial scheme follows the outlined method in the methodology section (Figure \ref{fig:cutequal}). The second scheme involves discretization from time zero to the maximum time point in the training set, with an additional time point corresponding to the event time (Figure \ref{fig:cutsame}). A comparative assessment of these two schemes was conducted across four numbers of time points, i.e., 3, 5, 10, 30, 50, and 100. The hyperparameters for ICTSurF for all simulations were similar to the baseline experiments. However, the number of nodes in each hidden layer and the dimension of time embedded vector were set to 64 and 16, respectively. For this experiment, we only used METABRIC and SUPPORT datasets.

\section{Results and Discussion}

Our study introduces a novel Deep Neural Network (DNN) model for survival analysis, comparing it with baseline models from previous DNN studies in survival analysis and the Cox Proportional Hazards (CPH) model as a baseline for traditional methods. We conduct comparative analyses on both single-risk datasets and competing-risk datasets. In addition, we investigate the effect of discretization schemes, which are necessary when using DNNs based on non-parametric approaches \cite{kvamme_continuous_2021}. Furthermore, we compare the effects of different time covariate representations, including raw time, time embedding with positional encoding, and time embedding with a learnable encoder.

We selected several DNN models for survival analysis, including SurvTRACE, DSM, DeepHit, and PC-Hazard, each chosen for specific reasons. PC-Hazard and SurvTRACE were chosen because their loss functions are based on the continuous-time survival model and hazard function, similar to our proposed model. SurvTRACE, in particular, is a more recent model that employs a more advanced architecture, a transformer, which we believe provides a strong basis for comparison with our model. On the other hand, PC-Hazard has a similar architecture to ours, making it a suitable baseline for comparison. We included DSM in our comparison to assess its performance against parametric models, providing insights into the strengths and weaknesses of different modeling approaches. DeepHit was selected as a benchmark for models using discrete time loss functions based on PMF, allowing us to evaluate the effectiveness of our model in comparison to such methods.

In this study, we utilize both public and simulated datasets previously employed in survival analysis. The METABRIC dataset has been extensively utilized in various studies such as DeepSurv, DeepHit, DSM, PC-Hazard, SurvTRACE, and ISF. Across these studies, there is a consistent definition of the dataset, leading to similar reported performances. Regarding the SUPPORT dataset, it has been utilized in DeepSurv, DSM, PC-Hazard, and SurvTRACE. Notably, there are two versions of the SUPPORT dataset, differing in the number of features employed. The DeepSurv version utilizes 14 features, while the DSM version incorporates 30 features. Previous studies utilizing the DeepSurv version of SUPPORT (DeepSurv, PC-Hazard, and SurvTRACE) have reported similar performances, whereas the DSM version exhibits higher performance, likely due to the increased feature set. For the SYNTHETIC NONLINEAR dataset, only the DeepSurv study has previously utilized this simulation method. Finally, in the case of SYNTHETIC COMPETING, only one prior study \cite{ahmed_m_deep_2017} has experimented with this simulated dataset.

\input{single_risk_results_untrack}
\input{competing_risk_results_untrack}

\subsection{Single risk dataset}

The performance of different models on the METABRIC dataset was evaluated using $C^{td}$-index and Brier score, as summarized in Table \ref{tab:single_risk}. ICTSurF demonstrated superior discriminative performance across all percentiles of event time (25\%, 50\%, and 75\%), achieving scores of 0.711$\pm$0.023, 0.696$\pm$0.020, and 0.665$\pm$0.030, respectively. SurvTRACE also showed strong calibration ability, especially at the $50^{\text{th}}$ and $75^{\text{th}}$ percentiles for the Brier scores. As seen in Table \ref{tab:single_risk}, SurvTRACE showed the best discriminative performance in the SUPPORT dataset, particularly in the $C^{td}$-index, with top scores at all quantiles (0.833$\pm$0.016, 0.789$\pm$0.007, and 0.751$\pm$0.006). ICTSurF and PC-Hazard closely followed, showing competitive performance. In the synthetic nonlinear dataset (Table \ref{tab:single_risk}), ICTSurF placed second across all quantiles with regards to the $C^{td}$-index (0.614$\pm$0.010, 0.609$\pm$0.006 and 0.602$\pm$0.004). SurvTRACE consistently outperformed other models in both $C^{td}$-index and Brier score across all percentiles.

In comparing various time features, our experiments revealed that ICTSurF with learnable encoding (Time2Vec) outperforms both $\text{ICTSurF}_{\text{pe}}$ and $\text{ICTSurF}_{\text{woe}}$ in overall performance (Table \ref{tab:single_risk}). The results further indicate that $\text{ICTSurF}_{\text{pe}}$ and $\text{ICTSurF}_{\text{woe}}$ exhibit similar performance in the SUPPORT and METABRIC datasets, while in the SYNTHETIC NONLINEAR dataset, $\text{ICTSurF}_{\text{pe}}$ demonstrates superior performance compared to $\text{ICTSurF}_{\text{woe}}$.

In summary, our performance outcomes are also similar those of prior studies employing similar datasets. Discrepancies may arise due to the stochastic nature of experiments and variations in performance metrics, with some studies employing Harrell's $C$-index and others opting for $C^{td}$-index without IPCW.
SurvTRACE demonstrated robust performance, with ICTSurF as a strong competitor. This aligns with the findings from Brier score evaluations, where SurvTRACE exhibited superior calibration ability, followed closely by ICTSurF. As expected, the traditional CPH model significantly lagged behind, notably evident in both its $C^{td}$-index and Brier score performances. Moreover, in the comparison of ICTSurF with PC-Hazard, which shares similarities such as a continuous-time survival model (similar to SurvTRACE and ICTSurF) and a comparable neural network architecture (feed-forward network), ICTSurF demonstrated stronger overall performance. This suggests that the ICTSurF modeling approach has the potential to achieve favorable outcomes compared to alternative methods, and there is a possibility of further improvement by leveraging more advanced neural network architectures, such as transformers.

\subsection{Competing risk dataset}

The comparative performance of ICTSurF with the baseline models on the dataset involving competing risks is presented in Table \ref{tab:compet_risk}. In relation to the $C^{td}$-index, ICTSurF consistently surpasses cs-CPH, DeepHit, and DSM across various quantiles and risks. Additionally, regarding the Brier score, ICTSurF exhibits the most superior calibration ability, with the exception of the first quantile for Risk 2 where DSM performs better. In general, the model's ability to consistently outperform other benchmarks indicates its efficacy in handling diverse risk scenarios.

\subsection{Comparison of discretization schemes}

\begin{figure*}[!tb]
     \centering
     \includegraphics[width=1.0\textwidth]{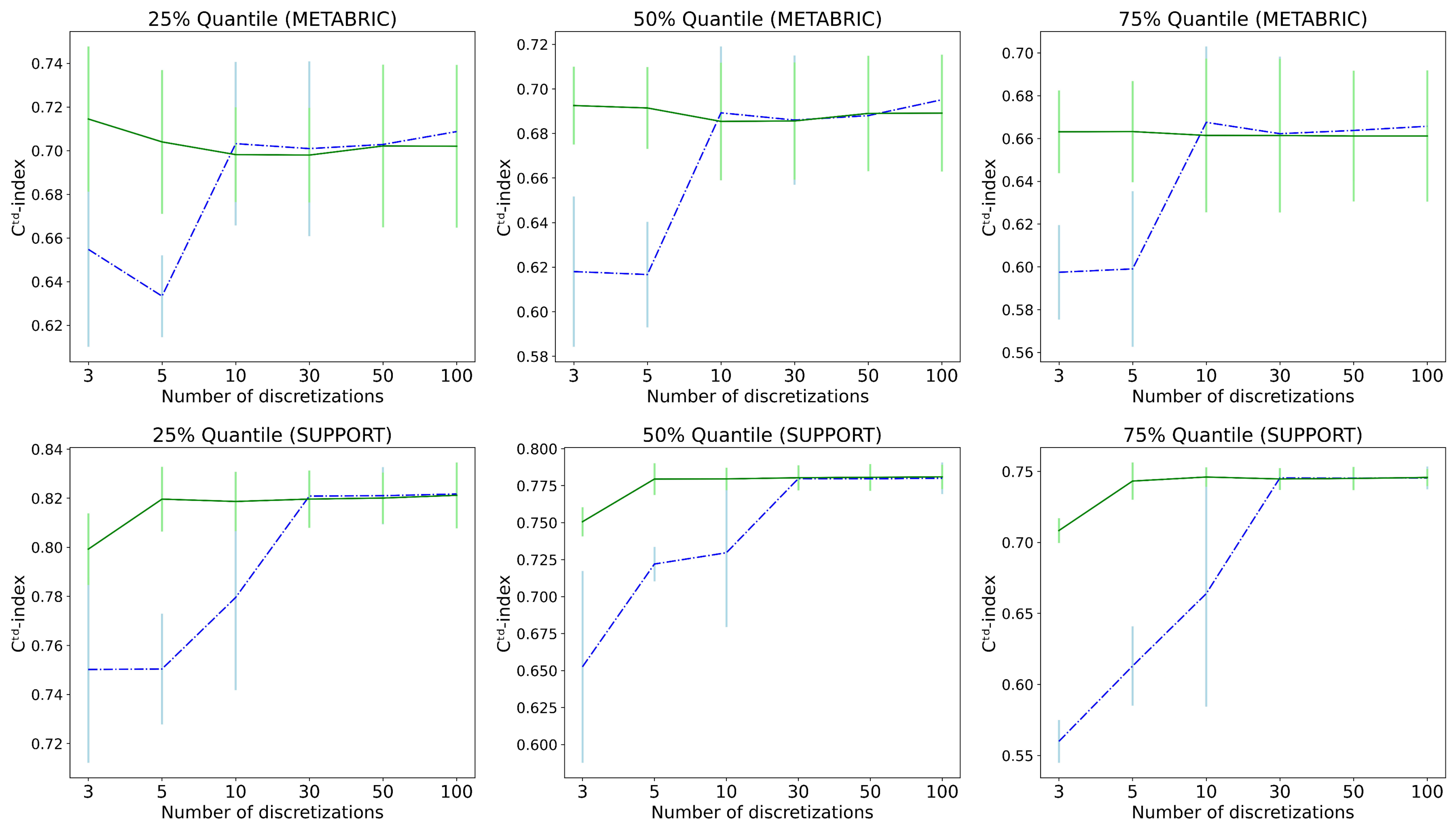}
     \caption{ Outcomes (mean and standard error of $C^{td}$-index) from Discretization Experiments at 25\%, 50\%, and 75\% event percentiles in METABRIC and SUPPORT datasets. The green solid line represents the discretization scheme as demonstrated in Figure \ref{fig:cutequal}, while the blue dashed line corresponds to the scenario illustrated in Figure \ref{fig:cutsame}.}
     \label{fig:compare}
\end{figure*}

In this comparison, we evaluated two discretization schemes, as illustrated in Figure \ref{fig:cuts}. In the first scheme, we leveraged the advantages of our method, while the second scheme attempts to replicate a common discretization approach, which uniformly divides the horizon and applies the same discretization to all samples.

The outcomes of these experiments are presented in Figure \ref{fig:compare}. For the METABRIC dataset, the first discretization scheme demonstrates consistent performance in terms of the $C^{td}$-index across all discretization numbers. On the contrary, the second discretization scheme demonstrates lower performance with a smaller number of discretizations, gradually improving to a level comparable to the first scheme as the number of discretizations increases, reaching stability at around 10 discretizations.

For the SUPPORT dataset, similar trends are observed, although it takes approximately 30 discretizations for the second scheme to achieve performance comparable to the first scheme.

In this section, we demonstrated that the flexibility of the discretization scheme enables us to utilize a lower number of discretizations. For instance, with the number of discretizations as low as three, the time horizon becomes {0, $T^*$, $T_{max}$} in the second discretization scheme for all samples, potentially leading to suboptimal performance. This aligns with findings from prior studies, such as \cite{ling_learning_2023, kvamme_continuous_2021}, where poor performance was observed with a reduced number of discretizations. 
Moreover, our method's ability to utilize different discretizations for distinct samples, as demonstrated in the first discretization scheme, contributes to enhanced computational efficiency compared to discrete-time survival model with implicit survival model. 

\subsection{Limitations}
One limitation is the availability and quality of public datasets for survival analysis, as they may not always accurately represent real-world clinical scenarios. Additionally, retrospective studies, common in survival analysis, may not always suggest generalizability to prospective data. Prospective evaluation is needed in future studies to assess this aspect. Another limitation is that our competing risk experiment relied on simulated data, which, while useful for validation and refinement, may not fully capture the complexity and variability of real-world data.
Lastly, our study focused on survival analysis without considering time-varying data. As a result, our model was not developed to retain memory of previous time steps, unlike models such as DynamicDeephit that incorporate recurrent neural networks. Future studies could aim to improve the model to accommodate time-varying data and enhance its performance in capturing temporal dependencies.


\section{Conclusion and Future Work}
In conclusion, ICTSurF showcases strong performance relative to existing methods in survival analysis. Its use of a continuous-time survival model enables a flexible discretization scheme for inputs and facilitates the modeling of interactions between covariates and specific times through the network architecture. Additionally, the incorporation of a learnable embedding layer for time information has been shown to enhance the model's performance. The implementation of ICTSurF is publicly available at https://github.com/44REAM/ICTSurF.

Future work involves extending the application of our method to complex datasets, such as those featuring time-varying covariates or different modalities, including image data. Additionally, there is a potential for experimentation with more modern neural network architectures to further enhance the capabilities and performance of our approach.

\section*{Declaration of Generative AI and AI-assisted technologies in the writing process}

During the preparation of this work the author(s) used ChatGPT in order to check grammar and wording of the manuscript. After using this tool/service, the author(s) reviewed and edited the content as needed and take(s) full responsibility for the content of the publication.

\bibliographystyle{elsarticle-num_untrack}
\bibliography{references_untrack}
\end{document}

%% file: dataset_untrack.tex
\begin{table*}[!tb]
\caption{Descriptive statistics of single risk and competing risk datasets.}
\setlength{\tabcolsep}{3pt}
\centering
\resizebox{16cm}{!}{
\begin{tabular}{|c|c|c|c|c|c|}

\hline
\multirow{2}{*}{Dataset} & \multirow{2}{*}{Dataset Type} & \multirow{2}{*}{Total Data}
& Feature Dim. & \multirow{2}{*}{No. Events} & \multirow{2}{*}{No. Censored} \\
& & & (real, categorical) & & \\
\hline
METABRIC
&Single risk&1,981 & 24 (16, 6)
&888 (44.83\%)&1,093 (55.17\%) \\

SUPPORT
&Single risk&9,105 & 28 (20, 8)
&6,201 (68.11\%)&2,904 (31.89\%) \\

SYNTHETIC NONLINEAR
&Single risk& 5,000 & 10 (10, 0)
&2,500 (50\%)&2,500 (50\%) \\

\multirow{2}{*}{SYNTHETIC COMPETING} & \multirow{2}{*}{Competing risk}
& \multirow{2}{*}{30,000} & \multirow{2}{*}{20 (20, 0)} &  Event 1: 6,846 (22.82\%) &  \multirow{2}{*}{15,000 (50\%)}\\ 
& &  &  &  Event 2: 8,154 (27.18\%)& \\

\hline
\end{tabular}
}
\label{tab:datasets}
\end{table*}

%% file: single_risk_results_untrack.tex
\begin{table*}[!tb]

\caption{$C^{td}$-index and Brier score for Single risk datasets}
\setlength{\tabcolsep}{3pt}
\centering

\resizebox{16cm}{!}{
\begin{tabular}{|c|c|c|c|c|c|c|c|c|c|}
\hline
\multirow{3}{*}{Models}& \multicolumn{9}{|c|}{C-index} \\
\cline{2-10}
& \multicolumn{3}{|c|}{METABRIC} 
& \multicolumn{3}{|c|}{SUPPORT} 
& \multicolumn{3}{|c|}{SYNTHETIC NONLINEAR} \\
\cline{2-10}
& 25\% & 50\% &75\%& 25\% & 50\% &75\%& 25\% & 50\% &75\% \\
\hline
CPH
&0.654$\pm$0.035&0.660$\pm$0.025 & 0.653$\pm$0.020
&0.800$\pm$0.014&0.764$\pm$0.009 & 0.740$\pm$0.006
&0.496$\pm$0.009&0.499$\pm$0.009 & 0.496$\pm$0.013  \\

PC-Hazard&0.684$\pm$0.027&0.685$\pm$0.023 & 0.656$\pm$0.030
&0.826$\pm$0.014&0.782$\pm$0.007 & \textit{0.747$\pm$0.006}
&0.523$\pm$0.013&0.517$\pm$0.013 & 0.516$\pm$0.013\\

DeepHit
&\textit{0.710$\pm$0.037}&0.688$\pm$0.030 & 0.571$\pm$0.030
&\textit{0.829$\pm$0.014}&0.768$\pm$0.012 & 0.638$\pm$0.018
&0.595$\pm$0.070&0.594$\pm$0.040 & 0.583$\pm$0.061 \\

DSM
&0.698$\pm$0.012&\textit{0.696$\pm$0.022} & 0.655$\pm$0.033
&0.801$\pm$0.013&0.761$\pm$0.004 & 0.711$\pm$0.013
&0.595$\pm$0.035&0.590$\pm$0.029 & 0.585$\pm$0.022 \\

SurvTRACE
&0.695$\pm$0.040&0.691$\pm$0.029&\textit{0.662$\pm$0.026}
&\textbf{0.833$\pm$0.016}&\textbf{0.789$\pm$0.007} & \textbf{0.751$\pm$0.006}
&\textbf{0.628$\pm$0.014}&\textbf{0.618$\pm$0.011} & \textbf{0.610$\pm$0.006} \\
\hline

$ \text{ICTSurF}_{\text{woe}} $
&0.698$\pm$0.025&0.678$\pm$0.018& 0.657$\pm$0.025
&0.809$\pm$0.010&0.772$\pm$0.009 & 0.742$\pm$0.007
&0.570$\pm$0.036&0.561$\pm$0.039 & 0.555$\pm$0.039 \\

$ \text{ICTSurF}_{\text{pe}} $
&0.696$\pm$0.030&0.678$\pm$0.025& 0.657$\pm$0.029
&0.816$\pm$0.017&0.769$\pm$0.009 & 0.721$\pm$0.012
&0.588$\pm$0.016&0.580$\pm$0.010 & 0.574$\pm$0.015 \\

 ICTSurF 
&\textbf{0.711$\pm$0.023}&\textbf{0.696$\pm$0.020}& \textbf{0.665$\pm$0.030}
&0.825$\pm$0.014&\textit{0.783$\pm$0.008} & 0.746$\pm$0.007
&\textit{0.614$\pm$0.010}&\textit{0.609$\pm$0.006} & \textit{0.602$\pm$0.004} \\
\hline
\multicolumn{4}{c}{}\\
\hline

\multirow{2}{*}{Models}& \multicolumn{9}{|c|}{Brier score}  \\
\cline{2-10}
& 25\% & 50\% &75\%& 25\% & 50\% &75\%& 25\% & 50\% &75\% \\
\hline
CPH
&0.099$\pm$0.010&0.167$\pm$0.010 & 0.207$\pm$0.009 
&\textbf{0.114$\pm$0.004}&0.168$\pm$0.006 & 0.189$\pm$0.004
&0.167$\pm$0.017&0.261$\pm$0.011 & 0.281$\pm$0.002 \\

PC-Hazard
&0.098$\pm$0.003&0.163$\pm$0.004 & \textit{0.206$\pm$0.011}
&0.127$\pm$0.007&\textit{0.159$\pm$0.004} & \textit{0.186$\pm$0.004}
&0.168$\pm$0.004&0.248$\pm$0.004 & 0.239$\pm$0.004\\

DeepHit
&0.098$\pm$0.002&0.173$\pm$0.004 & 0.231$\pm$0.006
& 0.124$\pm$0.002& 0.229$\pm$0.005 &0.287$\pm$0.011
&\textit{0.163$\pm$0.001}&0.240$\pm$0.001 & 0.232$\pm$0.006 \\

DSM
& \textit{0.098$\pm$0.001}& 0.166$\pm$0.006 &0.225$\pm$0.019
& 0.124$\pm$0.001& 0.177$\pm$0.002 &0.209$\pm$0.006
& 0.165$\pm$0.006& 0.239$\pm$0.008 &0.236$\pm$0.004 \\

SurvTRACE
&0.099$\pm$0.003&\textbf{0.161$\pm$0.007}&\textbf{0.202$\pm$0.019}
&0.133$\pm$0.007&\textbf{0.157$\pm$0.004} & \textbf{0.184$\pm$0.004}
&\textbf{0.159$\pm$0.003}&\textbf{0.229$\pm$0.005} & \textbf{0.218$\pm$0.002} \\

\hline

$ \text{ICTSurF}_{\text{woe}} $
&0.098$\pm$0.002&0.165$\pm$0.006& 0.213$\pm$0.014
&0.128$\pm$0.002&0.170$\pm$0.005 & 0.190$\pm$0.007
&0.167$\pm$0.003&0.252$\pm$0.006 & 0.263$\pm$0.010 \\

$ \text{ICTSurF}_{\text{pe}} $
&\textbf{0.097$\pm$0.003}&0.165$\pm$0.008& 0.211$\pm$0.014
&0.122$\pm$0.003&0.180$\pm$0.009 & 0.208$\pm$0.017
&0.163$\pm$0.003&\textit{0.237$\pm$0.002} & \textit{0.230$\pm$0.003} \\

 ICTSurF 
&\textbf{0.097$\pm$0.003}&\textit{0.162$\pm$0.009}&0.208$\pm$0.015
&\textit{0.118$\pm$0.001}&\textit{0.159$\pm$0.004} & 0.186$\pm$0.006
&0.164$\pm$0.003&0.248$\pm$0.005 & 0.266$\pm$0.006 \\

\hline
\end{tabular}}
\label{tab:single_risk}
\end{table*}

%% file: competing_risk_results_untrack.tex
\begin{table*}[!tb]
\caption{$C^{td}$-index and Brier score for Competing risk datasets}
\setlength{\tabcolsep}{3pt}
\centering
\begin{tabular}{|c|c|c|c|c|c|c|}
\hline
\multirow{3}{*}{Models}& \multicolumn{6}{|c|}{C-index}  \\
\cline{2-7}
& \multicolumn{3}{|c|}{risk 1} &\multicolumn{3}{|c|}{risk 2} \\
\cline{2-7}
& 25\% & 50\% &75\%& 25\% & 50\% &75\% \\
\hline
cs-CPH
&0.510$\pm$0.001&0.510$\pm$0.001 & 0.510$\pm$0.001 
&0.570$\pm$0.006&0.570$\pm$0.006 & 0.570$\pm$0.006  \\

DeepHit&0.660$\pm$0.011&0.660$\pm$0.011 & 0.660$\pm$0.011
&0.744$\pm$0.002&0.744$\pm$0.002 & 0.744$\pm$0.002 \\

DSM&0.711$\pm$0.007&0.712$\pm$0.006 & 0.703$\pm$0.010
&0.718$\pm$0.004&0.722$\pm$0.005 & 0.722$\pm$0.006 \\

\hline

$ \text{ICTSurF}_{\text{woe}}$
&0.714$\pm$0.005&0.714$\pm$0.005 & 0.714$\pm$0.005
&\textbf{0.762$\pm$0.005}&\textbf{0.762$\pm$0.005} & \textbf{0.761$\pm$0.005} \\

$ \text{ICTSurF}_{\text{pe}}$
&\textbf{0.717$\pm$0.004}&\textbf{0.717$\pm$0.004} & \textbf{0.718$\pm$0.004}
&\textbf{0.762$\pm$0.005}&0.762$\pm$0.006 & 0.760$\pm$0.006 \\

ICTSurF
&0.714$\pm$0.004&0.715$\pm$0.004 & 0.716$\pm$0.005
&0.759$\pm$0.004&0.758$\pm$0.004 & 0.755$\pm$0.004 \\

\hline
\multicolumn{4}{c}{}\\
\hline
\multirow{2}{*}{Models}& \multicolumn{6}{|c|}{Brier score}  \\
\cline{2-7}
& 25\% & 50\% &75\%& 25\% & 50\% &75\% \\
\hline
cs-CPH
&0.125$\pm$0.005&0.240$\pm$0.005&0.306$\pm$0.002 
&0.129$\pm$0.005&0.231$\pm$0.004&0.295$\pm$0.002  \\

DeepHit&0.178$\pm$0.002&0.205$\pm$0.001 & 0.249$\pm$0.002
&0.205$\pm$0.007&0.205$\pm$0.004 & 0.214$\pm$0.002\\

DSM&0.139$\pm$0.002&0.259$\pm$0.011 & 0.467$\pm$0.010
&\textbf{0.115$\pm$0.002}&0.181$\pm$0.002 & 0.174$\pm$0.006 \\

\hline

$ \text{ICTSurF}_{\text{woe}}$
&0.113$\pm$0.002&\textbf{0.196$\pm$0.001} & 0.175$\pm$0.003
&0.118$\pm$0.002&0.182$\pm$0.003 & 0.169$\pm$0.005 \\

$ \text{ICTSurF}_{\text{pe}}$
&0.113$\pm$0.002&0.198$\pm$0.001 & \textbf{0.175$\pm$0.002}
&0.118$\pm$0.002&0.182$\pm$0.002 & \textbf{0.168$\pm$0.005} \\

ICTSurF& \textbf{0.113$\pm$0.001}&0.198$\pm$0.002 & 0.175$\pm$0.003
&0.118$\pm$0.002&\textbf{0.180$\pm$0.003} & 0.171$\pm$0.005 \\

\hline
\end{tabular}
\label{tab:compet_risk}
\end{table*}